# Drone LAMS: A Drone-based Face Detection Dataset with Large Angles and Many Scenarios

Yi Luo[1], Siyi Chen[2], and X.-G. Ma[2]

[1] School of Energy and Environment, Southeast University, Nanjing 211100, China
[2] International Institute for Urban Systems Engineering, Southeast University, Nanjing 211100, China

Corresponding author: X.-G. Ma (e-mail: tiger_ma@seu.edu.cn).

This work is supported by the "SEU Double First-Class" University Project under Grant 4005002071, National Key Research and Development Project under Grant 6305001037.

**ABSTRACT** This work presented a new Drone-based face detection dataset. Drone LAMS solved low-performance issues of Drone-based face detection in scenarios such as large angles, which was a predominant working condition when a Drone flies high. The proposed dataset captured images from 261 videos with over 43k annotations and 4.0k images with pitch or yaw angle in the range of -90° to 90°. Drone LAMS showed significant improvement over currently available Drone-based face detection datasets regarding detection performance, especially with large pitch and yaw angles. Detailed analysis of how key factors, such as image similarity, annotation method，etc., impact dataset performance was also provided to facilitate further usage of a Drone on face detection.

**INDEX TERMS** face detection, image similarity, image dataset, Drone, UAV

## I. INTRODUCTION

Drones or UAVs were aircraft without any pilot on board and can be controlled remotely[1]. Drones with a pre-programmed mission could quickly get into inaccessible areas for human beings to perform various tasks where reduction of manpower consumption and material resources, and boost of productivity could be realized. Due to its large field view in the sky, drones are widely used in various areas such as forestry research[2], traffic motion route analysis[3], surveillance or monitoring [4, 5], aerial delivery system[6], and especially in various scenarios where object detection was the primary task, such as people and vehicle detection [7], human head counting [8], and people re-identification [9] widely used Drones. To fulfill the aforementioned tasks in real unconstrained condition, it is extremely important for the drone to have a robust object detection efficiency in the sky.

Although the advance of using a Drone to detect and monitor people was noticeable, very little research was available on Drone-based face detection and face recognition, limiting the application of Drones in real scenarios. Face recognition developed from traditional hand-craft feature methods to deep learning techniques. Usually, it consisted of following four steps: face detection, face alignment, face representation, and face matching [10]. Face detection's effectiveness could significantly influence the input quality for a face recognition system and eventually determine its overall performance. When a Drone flies high in the air, the face underneath to be detected could have a large pose, big angle, and many other harsh conditions, making the face detection more possible to lose its effectiveness.

Face recognition systems was high dimensional in order to shrink intra-class distance and enlarge inter-class distance [11] and dimensionality reduction was often needed [12], while face detection dataset was usually used for a binary classification task requiring large inter-class discrepancy only. Drone-based datasets currently available usually took both face detection and face recognition into consideration, and had low efficiency in training face detectors, and could not meet the requirement of researchers who need face detection systems with high efficiency. In order to improve the performance of face detectors, we need to evaluate training on Drone-based datasets and study its corresponding performance of face detectors with various distance, face pose, et al. In this work, we presented a Drone-based face detection dataset, Drone LAMS with significant improvement over currently available Drone-based face detection datasets regarding detection performance, especially with large pitch and yaw angles, aiming to boost research in drone-based face detection related areas.

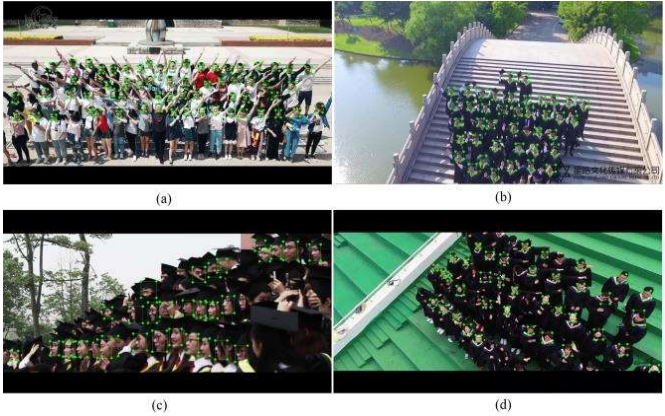

**FIGURE 1.** Typical images from Drone LAMS:
(a) Large pitch angle, slight yaw angle, sunny background;
(b) Large pitch angle, medium yaw angle, bright background;
(c) slight pitch angle, large yaw angle, cloudy background;
(d) Large pitch angle, large yaw angle, gray background.

*A. FACE DETECTION INTRODUCTION*

Face detection was an essential procedure for many facial applications, such as face alignment [13], face recognition [14], and face verification [11]. As the pioneer for face detection, Viola-Jones [15] adopted the AdaBoost algorithm with hand-crafted features, mainly relying on researchers' experience. After the famous face detection benchmark, WIDER FACE dataset was promoted[16], hand-crafted feature method was replaced by various deep learning methods. MTCNN [17] acquired high performance that hand-crafted features methods could never achieve before. As a face was also one kind of object, many face detectors benefited from the generic object detection methods[18-21]. Researchers conducted comprehensive experiments to study effects of small faces on detection performance[19, 20, 22, 23]. RetinaFace [19], based on RetinaNet [18], manually added five facial landmarks labels in WIDER FACE and introduced multitask loss as extra supervision. SSD-based $S^3FD$ [20] promoted a scale-equitable framework to try to solve the small face issues. DSFD [22] conducted a dual-shot face detector using Progressive Anchor Loss (PAL) and Improved Anchor Matching (IAM) strategy. Pyramidbox [23] accepted the same framework with $S^3FD$, and proposed Low-level Feature Pyramid Networks (LFPN) and Pyramid Anchors (PA) to fully use both lower-level features and high-level features.

In the work of Viola-jones [15], Jain[24], Mikolajczy [25], and Kienzle[26], face detectors were unable to perform on multi-pose, and FDDB [24] was limited to test only frontal face detection. Yang et al., [16] promoted the WIDER FACE benchmark and quantified its properties on scale, occlusion, pose, background clutters. Although some CNN-based face detectors [20, 22] tried to study the effects of small faces on detection performance, face detectors remained a challenge to capture faces with high degree of variability of face angle [16], especially in real-world scenarios, such as Drone-based face detection [27].

*B. EXISTING GENERIC FACE DETECTION DATASETS*

The face detection dataset could significantly affect a face detector's performance, especially for face detectors with the deep learning method [16]. Vidit and Erik's [24] had their dataset collected from the Yahoo website which contained many similar images in a real unconstrained world. Since face detectors could not benefit from pictures with the same features, Vidit and Erik removed duplicated images as many as possible from their collection to handle this issue. This made their dataset FDDB one of the most widely used face detection datasets.

According to the previous investigation into face detection datasets [16, 24], face detectors' efficiency was affected by the degree of variability in face pose, scenario, and face numbers, regardless of how the images were taken. Therefore, we chose to use generic face detection datasets such as WIDER FACE [16], FDDB [24], AFW [28], and PASCAL FACE [29], with a high degree of variability to compare with Drone-based face detection datasets for training and testing.

FDDB used to be the most comprehensive face detection dataset when it was posted in 2017 with 2,845 images and 5,171 annotated face boundary boxes included, and was the first one having multiple faces in one image.

WIDER FACE was ten times larger than IJB-A [30] and had 393,703 labeled face boundary boxes in 32,203 images with a higher degree of variability in face pose, scenario, occlusion, and ambiguity over FDDB. Because datasets built before WIDER FACE had limited variability and faces, face detectors could be easily saturated on these face detection datasets and could not give high performance. For instance, before WIDER FACE was promoted, the best average precision (AP) performance on AFW and PASCAL FACE was 97.2% and 92.11%, respectively, and the highest Recall on FDDB was 91.74%. However, the Recall of the easy, medium, hard portion of the WIDER FACE could only reach 92%, 76%, and 34% tested by EdgeBox [31]. In this work, we chose WIDER FACE as the representative of generic face detection dataset for comparison.

TABLE I
DATASETS FOR TRAINING

| Face dataset | Range of face pitch angle | Subject | DR | Annotated face | Scenario |
|---|---|---|---|---|---|
| WIDER FACE | (-90, 90) | 393k | <0.01% | 393k | 1.3w |
| Drone FACE | (-60, 0) | 11 | 9.09% | None | 2 |
| IJB-S | (-20, 20) | 202 | 0.49% | None | 10 |
| Drone SURF | (-30, 30) | 58 | 1.72% | 786k | 1 |
| Drone LAMS | (-90, 90) | 10k | <0.05% | 43,531 | 261 |

Even though WIDER FACE provided photos with high variability in face pose, scenario, etc., the images in WIDER FACE mainly gathered from frontal view of faces. However, images were usually collected underneath a drone and had a relatively large angle in drone-based face detection scenarios. In other words, the performance of drone-based face detectors could be negatively impacted if general face detection datasets were used directly to train drone-based face detectors.

### C. DRONE-BASED FACE DETECTION DATASETS

Drone-based face detections brought a set of challenges such as high degree of variability in face pose, illumination, and grave imbalance between positive and negative samples [27]. In contrast to general face detectors, where face is collected in front of a camera, face is usually underneath a Drone's camera, sometimes with very large angle. To deal with the challenges on drone-based face detection, a dataset with a large range of face pose, illumination, ambiguity, and capacity for training is highly needed.

Table I listed several datasets for face detection currently available. DRONE FACE used a stationary GoPro camera to collect images, and had 620 raw images of 11 people with distance and height annotation on an unconstrained scenario. The DRONE FACE did not provide the annotation of face bounding box, and cannot be used for training directly. IJB-S [32] had 10 drone-based videos of 202 people without annotation, and 350 surveillance videos and 202 enrollment videos with over 10 million annotations. DRONE-SURF [33] used drones to collect 200 videos out of 58 people with 786k annotated face bounding boxes under two scenarios. However, the images extracted from the videos had high image similarity.

## II. Drone LAMS
### A. OVERVIEW

Drone-based face detection datasets, such as Drone FACE, Drone SURF, and IJB-S, conducted experiments with limited people involved. A small number of people in the experiments caused high image similarity. We used the perceptual image hash function [34] with Discrete Cosine Transform (DCT) based Hash to calculate Hamming distance [35]. Type-Ⅱ DCT in Formula 1 was used in this work.

$$X[n] = \sqrt{\frac{2}{N}} \sum_{m=0}^{N-1} f[m] \cos(\frac{(2m+1)n\pi}{2N}) \quad (1)$$
$$, (n = 0, ..., N-1)$$

where f[m], m = 0, …, N - 1 represented pixel sequence of images. To quantify the face identity duplication, we used the Duplication Rate (DR), as described in Formula 2.

$$DR = \frac{1}{n} \sum_{i=1}^{n} \frac{M_i}{N} \quad (2)$$

where n represented numbers of people tested in a dataset, N represented the image numbers of a dataset, and $M_i$ represented numbers of a person's face used in a dataset.

Table I listed several key parameters of Drone-based datasets. The annotated face number was listed in Table I to represent the comprehensiveness and complexity of a dataset. When the DR is same, more annotated face numbers would provide higher detection performance. The number of scenarios was also listed since it would directly influence the generalization performance of a dataset. With wider difference in scenarios, the face detector training on the dataset will have more robust performance.

In order to improve detection performance, researchers often trained the object detectors offline and adapted data augmentation method to enrich the variability of input images. However, the effect of face data augmentation was hard to quantify [36], and the induced ambiguity would lead to unaccepted results. For video-based object datasets, frames were relatively small in variation, and only frames with large variation were selected for training and testing portion. When the captured videos were limited in variation such as scenarios, subject, and face pose, et al., the data augmentation method could have little effect on improvement of detection performance. Correia et al. [36] used data augmentation methods to enlarge the datasets with different face identities to make DR smaller. To some extent, Correia's method enlarged the dataset's capacity and therefore overcame the shortcoming of training a dataset with high DR. However, all the enlarged samples had the same face pose. Face detection could introduce object detection approaches, such as photometric distortions and geometric distortions [37], to further enrich training data.

DRONE LAMS adapted web crawler method to gather 527 drone-based videos of various scenarios such as, graduation ceremonies, sport games, etc., from the internet to overcome high duplication and other issues. We selected 261 videos which had high degrees of variability in scenarios, especially with pitch angle ranged from nodding (negative pitch angle) to up-looking (positive pitch angle), and yaw angle ranged from people right (positive yaw angle) to people left (negative yaw angle) as shown in Figure.1. All videos were from www.skypixel.com using various devices including DJI drone, OSMO, etc. The dataset contained 4001 images with high degree of variability in face pose and scenario, and had over 43k face annotations. The percentage of indoor images was around 28%.

DRONE LAMS had images with pitch angle in the range of -90° to 90°, while DROEN FACE, DROEN SURF, IJB-S only contained faces with relatively small range of pitch angle. From the aspect of people involved, DRONE FACE, DROEN SURF, IJB-S conducted experiments with limited people. This caused higher Duplication Rate (DR), as described in formula.1 over general face detection dataset, such as WIDER FACE. Drone LAMS had low DR by introducing large amount of different videos rather than few people in experiments.

In order to cut down face duplication rate, we monitored each video and grabbed images with high degree of variability in face pose and scenario. To the best of our knowledge, Drone LAMS had the lowest face duplication rate in current available drone-based face detection dataset, and could be used to train face detectors of drones in various unconstrained scenarios.

### B. COLLECTION METHODOLOGY

Drone LAMS adapted a web crawler method to gather Drone-based videos of various scenarios. We collected the raw videos following four steps: 1) Manually defined the video events with keywords to ensure that searching engines could find videos with faces via Drones' camera; 2) Used web crawler tools via the Skypixel's API interface to collect videos with HDMI resolution of 1920 × 1080; 3) Manually cleaned the videos, filtered out videos with no human face, and selected videos with high degrees of variability in scenarios, e.g., with pitch angle ranged from nodding (negative pitch angle) to up-looking (positive pitch angle) and the yaw angle from people right (positive yaw angle) to people left (negative yaw angle); 4) Calculated each image's perceptual hash value and removed similar images with hamming distances smaller than 5 to lower DR.

Two criteria were used to capture the Drone LAMS images: 1) If the Drone's relevant pose to the face did not change and only the distance changed, i.e., only the face size changed, we saved the minimum or the maximum face size image for simplicity; 2) If a person had nearly the same face pose in a video, we randomly saved one picture.

### C. ANNOTATION METHOD

Considering the deep learning method in object detection, a model would have better performance if a highly accurate annotated bounding box is available. For Drone-based face detection datasets listed in Table I. Drone FACE and IJB-S did not have annotations. Boundary area of images in DRONE SURF was too wide that unrelated body parts such as shoulders were also included. This made the annotations ineffective and was harmful to face detectors. This is partially because that the dataset was also used for face recognition [33]. For face recognition, each people should have enough quantity of samples with variation, resulting in heavy duplication rate. In comparison with WIDER FACE, datasets with heavy duplication rate would lead to poor performance of face detection.

We used the labelme tool [44] to annotate bounding box of a face in Drone LAMS, where face boundaries tightly covered forehead, chin, and cheek of the face, as shown in Fig.1. Deng [19] added additional five landmarks, i.e. each center of eye, nose tip, and mouth corners, in WIDER FACE for training, and acquired 52.297% mAP on WIDER face and pedestrian challenge 2018 [45]. Vidit and Erik [24] used elliptical face annotation, which worked well when faces had small pose. But if a face had large pose and occlusion, labelling the face became tough. The elliptical face annotation method was hard to estimate on large angles, because the ellipse needs two focus, semimajor axis of ellipsoid and semi minor axis, all of which are hard to label. On the contrary, tight boundary box of face is easy to label, and can benefit from additional landmarks to promote face detector performance [16, 19, 24].

If a face was occluded and had over 50% area that could be recognized by people, we still labeled the face boundary box with prediction area of occlusion. Considering drone captured face images sometimes was relatively small, we solely labeled faces with width length and height length both larger than 15 pixels. In order to annotate bounding box precisely and reduce consumption of manpower, each annotation used RetinaFace face detector with Mobilenet backbone trained on WIDER FACE to create pre-annotation. Subsequently, a face was labeled by an annotator and was cross-checked by two different people, as similar to S. Yang's work [16].

### III. EXPERIMENTS
#### A. FACE DETECTOR

RetinaFace [19] inherited RetinaNet [18] merits, which uses focal loss to solve the imbalance between positive and negative classes. For face detection, positive class means a face, and negative class means background with no face. In Drone-based face detection, the imbalance issue is extremely predominant, especially when the drone flies high, resulting in very small faces in one image. High-resolution images, such as 4K, could be used to solve the imbalance challenge, but this would increase calculation complexity and power consumption of GPU and other hardware.

The images could be resized to smaller ones, such as 1080P, but this will challenge feature extraction ability of the algorithm, and would be easy to diverge. As RetinaFace used focal loss method to deal with the class imbalance problem and acquired convincing performance on WIDER FACE, we chose RetinaFace, which has the ability to extract features from 1080P resolution images, as face detector for this work.

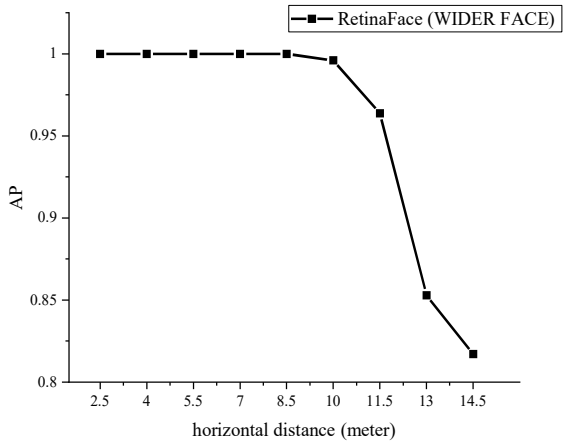

**FIGURE 2.** Horizontal AP performance of Drone FACE with RetinaFace trained on WIDER FACE, with yaw angle and pitch angle equal to 0°, and height of 1.5-meter.

For details, we used a light-weight Pytorch version RetinaFace with Mobilenet [46] backbone with a size of 1.8 Mb. This light-weight RetinaFace achieved 80.99% AP on WIDER FACE hard Val dataset.

We conducted all experiments on a workstation with an E5-2650 V3 CPU, 64G RAM, and 1080 Ti GPU in an Ubuntu 18.04 system.

We used brackets to represent the training dataset, i.e., RetinaFace (WIDER FACE) means RetinaFace trained on WIDER FACE. Although we used different datasets for training, the training method was the same, i.e., stochastic gradient decent (SGD) optimizer was used with 0.9 momentum, 0.0005 weight decay, and 64 batch size. The learning rate started from 0.001 and decreased to 0.01 after 5 epochs. After 200 epochs, the training process was terminated.

Average Precision (AP) was a single value, defined as the area under the Precision-Recall (P-R) curve, and was one of the standard indicators for information retrieval tasks [47]. We used AP in Formula 3, and mean Average Precision (mAP) in Formula 4 below to measure detectors' performance.

$$AP = \int_0^1 p(r)dr \quad (3)$$

$$mAP = \frac{1}{N}\sum_{i=1}^{N} AP_i \quad (4)$$

p represents Precision, r represents Recall, and N represents the class category.

### B. PERFORMANCE ALONG WITH HORIZONTAL AND VERTICAL DIRECTIONS

In a Drone-based face detection dataset, the pitch angle, yaw angle, and distance between the face and the Drone would influence the accuracy of face detectors in different manners, requiring to evaluate the variation separately.

We firstly controlled pitch angle and yaw angle at 0° and investigated into the effect of horizontal distance on AP. We used WIDER FACE as training dataset and DRONE FACE as the testing dataset. For details, we selected horizontal images in DRONE FACE, which had distance and height annotations, for calculating distance, pitch angle and yaw angle. Although DRONE FACE has no face boundary box, we annotated 620 raw images and used them for training.

Figure 2 showed that the AP value of RetinaFace was 100% with horizontal distance smaller than 8.5 meter. The AP decreased significantly when horizontal distance between a face and a camera was larger than 8.5 meter. This was caused by the fact that the area of the face in the image was smaller than the minimum receptive field of the face detector, and reached the bottleneck of the feature extraction ability.

In order to study the influence of pitch angle on AP, we controlled yaw angle at 0° and conducted experiments on a vertical plane, which is a combination of distance and pitch angle as shown in Figure.3.

In Fig.3, the AP value decreased significantly when the pitch angle or horizontal distance increased. It was also noticed that when the horizontal distance was larger than 14.5 meter, the AP value fluctuated significantly. This was largely because the RetinaFace reached its maximum ability of face detection at the 14.5 meter, resulting into irregular fluctuation.

In H.-J. Hsu's work [27], distance from the face to the drone was controlled within 4 meter, and only 65 images of DROEN FACE were used to test face detectors. In our experiments, RetinaFace acquired 100% within 4 meter, indicating that DROEN FACE did not have enough images to study the effect of pitch angle on AP with RetinaFace. In order to compare the AP performance of different datasets, we used datasets with more images and larger face pose on pitch angle and yaw angle to test face detectors.

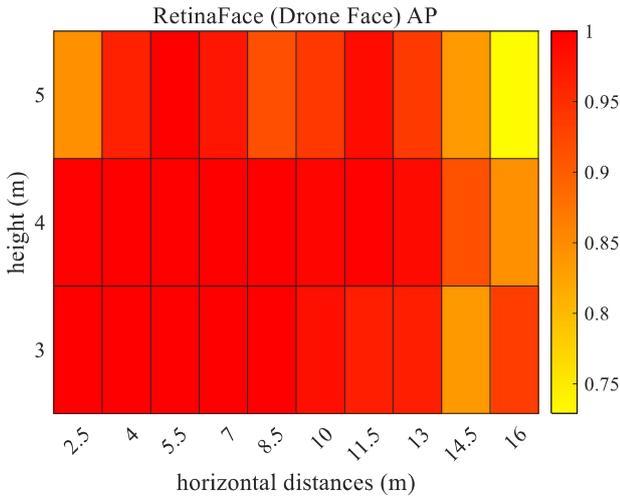

**FIGURE 3.** AP performance of Drone FACE with RetinaFace trained on WIDER FACE, with yaw angle equal to 0°.

### C. TESTING ON POINTING'04 AND AFLW

We adopted Pointing'04 dataset [48] and AFLW [49] to test the effect of the pitch angle and yaw angle on AP, since they had abundant face poses in images. Although Pointing'04 and AFLW were mainly used for face pose estimation, its large variation of angle attribute could also be used to study the effect of pitch and yaw angle on face detection performance. The Pointing'04 contained 2,790 images from 15 people in constrained world with variations of pitch and roll from -90 to +90 degrees, and each people had two image groups, each of which had 93 different face pose images. The AFLW had many images with more than one face and this might lead to wrong attribute of face angles. Thus, we selected images with only one face for this study. We chose 16,566 images from AFLW with pitch angle in the range of -60 to +60 and yaw angle in the range of - 90 to +90.

The label style of Pointing'04 was a fixed square boundary box with a size of 120×120 pixels that contained face and hair. The annotation style of the testing dataset was different from that of training datasets, such as WIDER FACE and Drone LAMS, and we changed the Pointing'04 annotation style to fit the annotation of DroneLAMS. Otherwise, the different label styles between training and testing datasets could cause inaccurate results.

Testing on Pointing'04 was shown in Figure 4 to study the influence of pitch angle on face detectors. Since AP distribution showed symmetrical results with pitch angle in the range of -90° to 90°, we only showed data from -90° to 0° for clarification.

In Fig.4, RetinaFace (DROEN FACE) showed the lowest AP performance, for DRONE FACE only contained frontal face images and images with pitch angle in the range of -60° to 0°. The minor face numbers was another reason for the weakest AP performance of RetinaFace (DRONE FACE). DRONE FACE also had shortcomings in slight scenario variation and higher DR comparing with general face detection datasets. The AP performance of RetinaFace (DRONE FACE) increased with decreasing pitch angle. This largely attributed to sample imbalanced distribution of the DRONE FACE dataset, where H.-J. Hsu [27] collected 97.58% of the face images with pitch angles in the range of -45° to 0°. Compared with RetinaFace (Drone LAMS) and RetinaFace (DRONE SURF), the AP performance of the models trained on DRONE FACE were not competent in real unconstrained scenarios.

RetinaFace (DRONE SURF) performed better than RetinaFace (DRONE FACE), mainly because it had more face images and lower DR over DRONE FACE. With rich samples of frontal faces in DRONE SURF, it obtained 100% AP with a pitch angle equal to 0°.

RetinaFace (Drone LAMS) obtained the highest AP performance for all pitch angles, as shown in Fig.4. As the Pointing'04 merely containing 11 people at 93 face pose images, the small testing dataset easily saturated according to the previous research [16, 24]. Drone LAMS performed better than DRONE FACE and DRONE SURF because it had the highest degree of variability in scenarios, face pose, and light.

For evaluating the influence of yaw angle to face detectors, we conducted an AP performance test of RetinaFace on Pointing'04 in Figure 5. The results showed similar tendency comparing to Fig.4, and the reason was that with the increasing of yaw angle, the difficulty of face detection also increased.

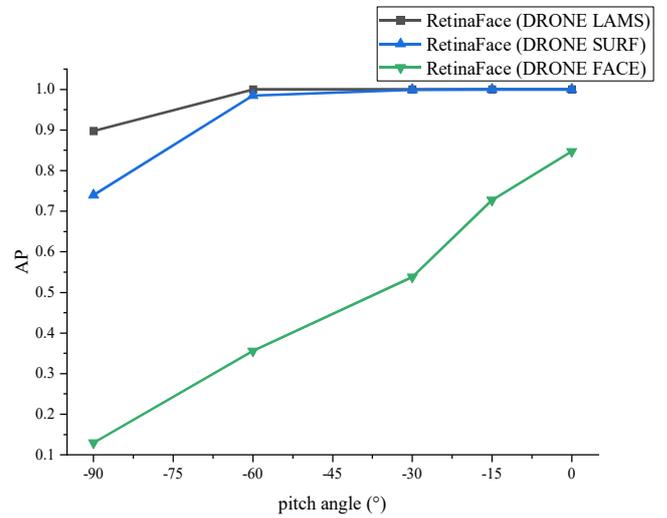

**FIGURE 4.** AP performance of Pointing'04 with RetinaFace trained on Drone LAMS (solid black square), DRONE SURF (blue triangle), and DRONE FACE (green triangle), respectively, with pitch angle in the range of -90° to 0°, yaw angle equal to 0°, and horizontal distance equal to 2 meter.

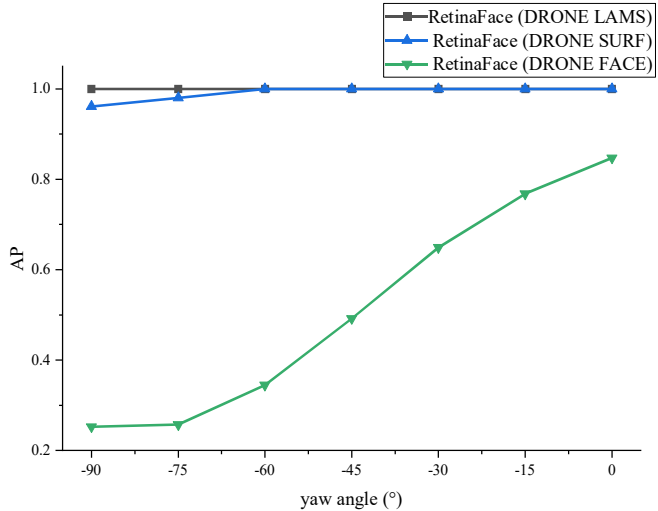

**FIGURE 5.** AP performance of Pointing'04 with RetinaFace trained on Drone LAMS (solid black square), DRONE SURF (blue triangle), and DRONE FACE (green triangle), respectively, with yaw angle in the range of -90° to 0°, pitch angle equal to 0°, and horizontal distance equal to 2 meter.

The overall AP performance was also tested on 93 different face postures in the Pointing'04 dataset, using all yaw angles and pitch angles images with distance equal to 2 meter, as shown in Table II. Drone LAMS had a noticeable improvement over DRONE SURF and DRONE FACE. RetinaFace trained on Drone LAMS had the highest generalization ability in Pointing'04. Compared to DRONE SURF, mAP trained on Drone LAMS was improved by 3.24% and nearly reached its maximum capability.

As Drone LAMS and Drone SURF both saturated on Pointing'04 for pitch angle and yaw angle in the range of -60° to +60°, Pointing'04 was unable to differentiate them preciously. AFLW was more comprehensive than Pointing'04, and always used for face pose estimation in real world. It had thousands of scenarios in the real unconstrained world with a high degree of variability in face poses and persons, and had more variation to evaluate the face detector with pitch angle in the range of -60° to +60°. In this work, we used pose annotation to evaluate face detector performance on different face postures to evaluate the pitch angle in the range of -60° to 60°. Since AP distribution showed symmetrical results with the pitch in the field of -60° to 60°, we only showed data from -60° to 0° for clarification.

In Figure 6, the AP value trained on Drone LAMS was the highest among the three datasets, and AFLW turned out to be more effective in evaluating the pitch performance of face detectors over the DRONE FACE and Pointing'04. In Table III, RetinaFace (DRONE SURF) had nearly 10% AP over RetinaFace (DRONE FACE), and RetinaFace (Drone LAMS) performed almost 50% better than RetinaFace (DRONE SURF). The results proved again that Drone LAMS has more comprehensiveness for researchers to train models over DRONE SURF and DRONE FACE.

## IV. CONCLUSION

In this work, we proposed Drone LAMS: a drone-based face detection dataset with large angles and many scenarios. 4001 images with high degree of variability in face pose and scenario, and over 43k face annotations were gathered from 261 drone-based videos. Drone LAMS showed convincing improvement of detection accuracy of drone-based face detectors, in both constrained and unconstrained scenarios. On Pointing'04 or AFLW, Drone LAMS outperformed other drone-based face detection datasets by up to 35% since it had the lowest face duplication rate among current available drone-based face detection datasets.

TABLE II
EVALUATION MAP OF SEVERAL DRONE-BASED FACE DETECTION DATASETS ON POING'04

| Face detector | mAP on Pointing'04 |
|---|---|
| RetinaFace (Drone LAMS) | 99.79% |
| RetinaFace (Drone SURF) | 96.27% |
| RetinaFace (Drone FACE) | 63.01% |

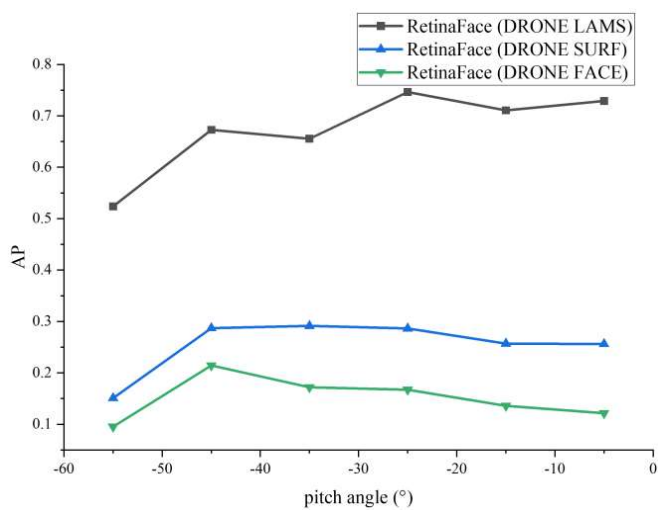

FIGURE 6. AP performance of AFLW with RetinaFace trained on Drone LAMS (black solid square), DRONE SURF (blue triangle), and DRONE FACE (green triangle), respectively, with pitch angle in the range of -90° to 0°, and yaw angle equal to 0°, and horizontal distance within 4 meter.

TABLE III
EVALUATION MAP OF SEVERAL DRONE-BASED FACE DETECTION DATASETS ON AFLW

| Face detector | mAP on AFLW |
|---|---|
| RetinaFace (Drone LAMS) | 73.64% |
| RetinaFace (Drone SURF) | 23.03% |
| RetinaFace (Drone FACE) | 11.98% |